\documentclass[10pt,twocolumn,letterpaper]{article}

\usepackage{cvpr}              

\usepackage{graphicx}
\usepackage{amsmath}
\usepackage{amssymb}
\usepackage{booktabs}
\usepackage{multirow}

\usepackage{algorithm}
\usepackage{algorithmic}

\usepackage[accsupp]{axessibility}

\usepackage[pagebackref,breaklinks,colorlinks]{hyperref}

\usepackage[capitalize]{cleveref}
\crefname{section}{Sec.}{Secs.}
\Crefname{section}{Section}{Sections}
\Crefname{table}{Table}{Tables}
\crefname{table}{Tab.}{Tabs.}

\def\cvprPaperID{7536} 
\def\confName{CVPR}
\def\confYear{2022}

\begin{document}

\title{Hyperspherical Consistency Regularization}

\author{Cheng Tan$^{1,2,*}$, Zhangyang Gao$^{1,2,*}$, Lirong Wu$^{1,2}$, Siyuan Li$^{1,2}$, Stan Z. Li$^{1,2}$ \\
$^{1}$ AI Lab, School of Engineering, Westlake University \\
$^{2}$ Institute of Advanced Technology, Westlake Institute for Advanced Study \\ \thanks{Equal contribution}
{\tt\small \{tancheng,gaozhangyang,wulirong,lisiyuan,stan.zq.li\}@westlake.edu.cn}
}
\maketitle
\begin{abstract}
Recent advances in contrastive learning have enlightened diverse applications across various semi-supervised fields. Jointly training supervised learning and unsupervised learning with a shared feature encoder becomes a common scheme. Though it benefits from taking advantage of both feature-dependent information from self-supervised learning and label-dependent information from supervised learning, this scheme remains suffering from bias of the classifier. In this work, we systematically explore the relationship between self-supervised learning and supervised learning, and study how self-supervised learning helps robust data-efficient deep learning. We propose \textit{hyperspherical consistency regularization} (HCR), a simple yet effective plug-and-play method, to regularize the classifier using feature-dependent information and thus avoid bias from labels. Specifically, HCR first project logits from the classifier and feature projections from the projection head on the respective hypersphere, then it enforces data points on hyperspheres to have similar structures by minimizing binary cross entropy of pairwise distances' similarity metrics. Extensive experiments on semi-supervised and weakly-supervised learning demonstrate the effectiveness of our method, by showing superior performance with HCR.
\end{abstract}

\section{Introduction}
\label{sec:intro}
The last decade has witnessed revolutionary advances in deep learning across various computer vision fields such as image classification~\cite{krizhevsky2012imagenet, he2016deep, huang2017densely,tan2019efficientnet}, object detection~\cite{ren2015faster,liu2016ssd,redmon2016you,redmon2017yolo9000}, and semantic segmentation~\cite{he2017mask,lin2017feature,ronneberger2015u} in the presence of large-scale labeled datasets. However, massive collection and accurate annotation of datasets are time-consuming and expensive. In many practical situations, only small-scale high-quality labeled datasets are available. For this reason, semi-supervised learning (SSL) that learning from few labeled data and a large number of unlabeled data has received broad attention~\cite{rasmus2015semi,tarvainen2017mean,berthelot2019mixmatch,berthelot2019remixmatch,li2019dividemix,tan2021co,xie2020unsupervised,sohn2020fixmatch,pham2021meta,xia2020part}.

\begin{figure}[ht]
  \centering
  \includegraphics[width=0.32\textwidth]{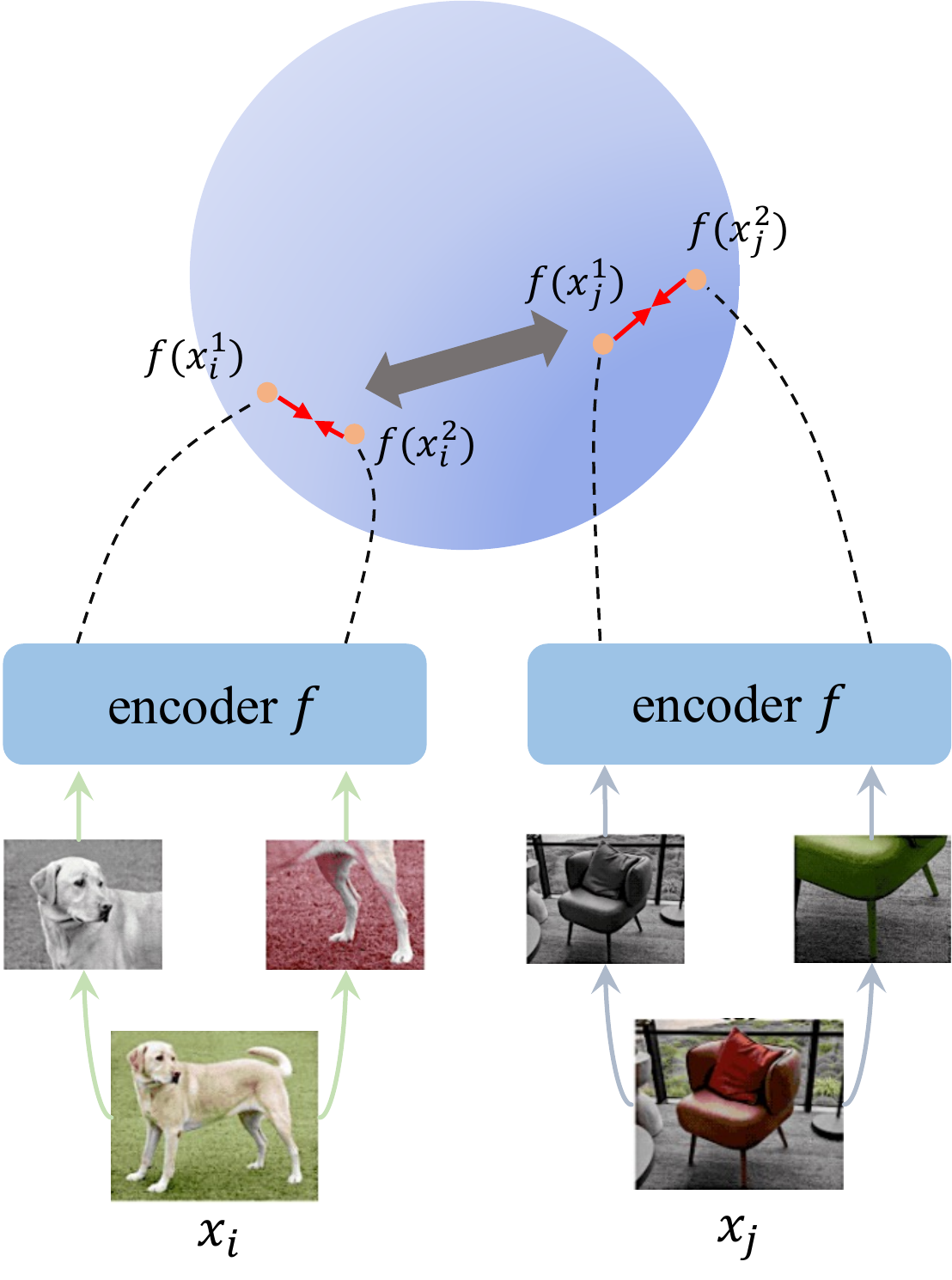}
  \caption{Illustration of contrastive learning on the hypersphere. Red arrows denote positive pairs tend to attract each other, and the gray arrow denotes negative pairs tend to repel each other.}
  \label{fig:cl_on_hypersphere}
\end{figure}

With the development of contrastive learning~\cite{he2020momentum,henaff2020data,chen2020simple,NEURIPS2020_f3ada80d,chen2020improved,chen2020big,caron2020unsupervised,li2020prototypical,wang2020understanding,zoph2020rethinking,hu2021adco,zbontar2021barlow,tianyuan_directpred}, recent SSL algorithms~\cite{wang2021self,zhong2020bi,li2020comatch,khosla2020supervised,han2020self,tack2020csi,wang2021contrastive} tend to extend self-supervised learning into supervised learning by adding a branch network as a projection head that jointly learns from feature-dependent and label-dependent information. Though the feature encoder is supposed to learn better by making agreements from different views on latent spaces, the classifier which determines the ultimate predictions still suffers from the bias of semi-supervision or weak-supervision. Typically, ~\cite{kang2019decoupling,zhou2020bbn} found that data imbalance is not the key issue in learning high-quality representations from long-tail data, while simply adjusting the classifier with balanced sampling can effectively alleviate the imblanced bias. This phenomenon suggests that decent representation may help but not be enough for robust learning, while regularizing the classifier is necessary to improve learning performance.

A vast number of current empirical contrastive learning methods~\cite{he2020momentum,chen2020simple,chen2020improved,NEURIPS2020_f3ada80d,chen2020big,li2020prototypical,hu2021adco} project feature embeddings on a hypersphere through $\ell_2$ normalization while maximizing distances between negative pairs and minimizing distances between positive pairs, as shown in Figure~\ref{fig:cl_on_hypersphere}. Restricting the output space to a unit hypersphere can improve training stability in machine learning where dot products are ubiquitous~\cite{wang2020understanding, xu2018spherical,wang2017normface}. Besides, well-clustered features on the hypersphere are linearly separable from the rest of the feature space. The above desirable traits are considered to be useful while regularizing the classifier.

\begin{figure}[ht]
  \centering
  \includegraphics[width=0.39\textwidth]{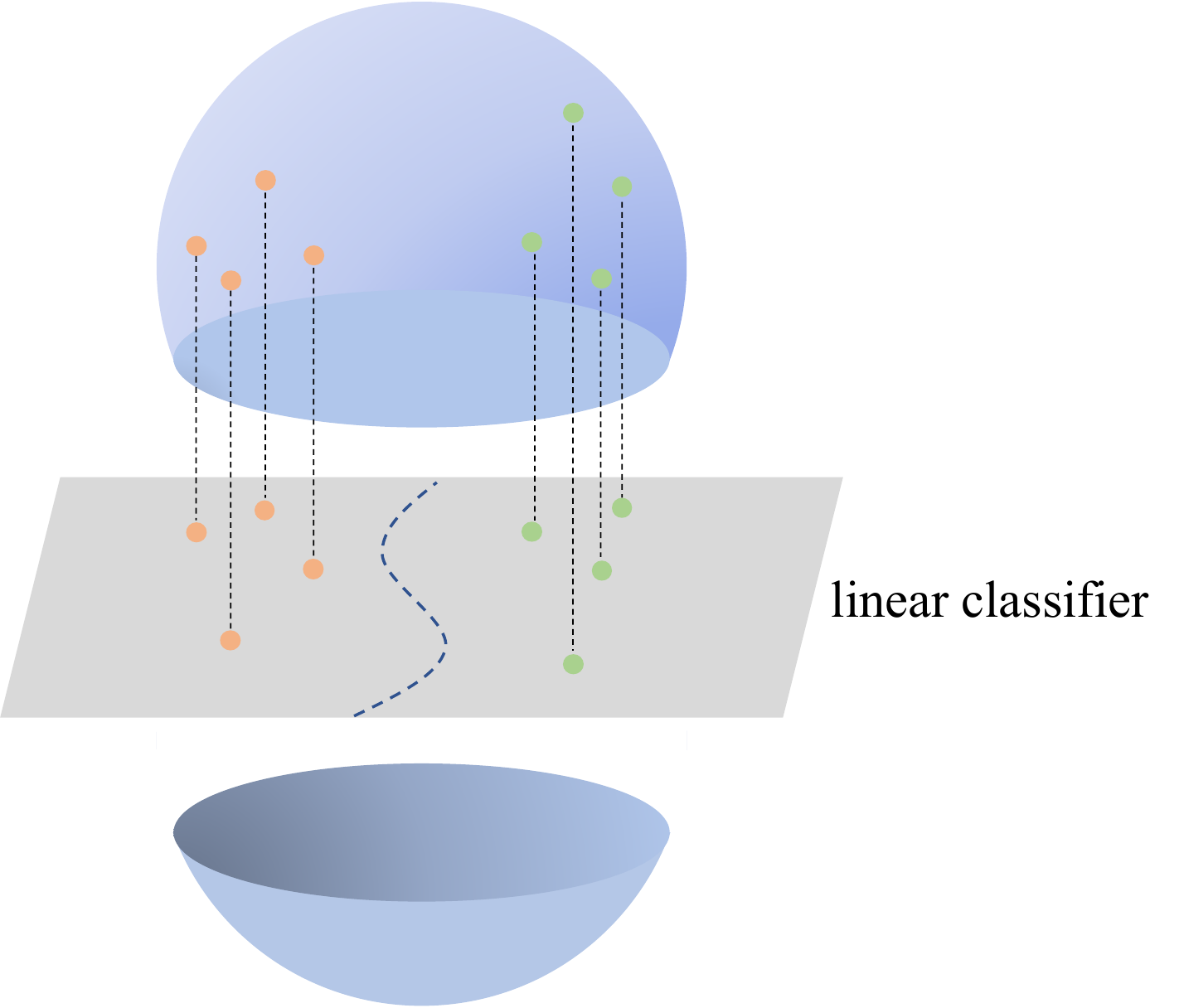}
  \caption{Linear classifier learns to separate the hypersphere through the hyperplane.}
  \label{fig:seperable_hypersphere}
\end{figure}

In this work, we analyze the relationship between the projection head and the classifier, and propose hyperspherical consistency regularization (HCR) to constrain the latent hyperspherical space. As shown in Figure~\ref{fig:seperable_hypersphere}, a decent classifier is able to find an optimal hyperplane in a hypersphere manifold, and data points on the classifier's hyperplane can be reprojected on a hypersphere. HCR assumes data points on the projection head's hypersphere and the classifier's hypersphere have similar geometric structures, and preserves such structures by making distributions of pairwise distances consistent. Experiments on semi-supervised learning and weakly-supervised learning indicate HCR can considerably improve the generalization ability.

\begin{figure*}[ht]
  \centering
  \includegraphics[width=\textwidth]{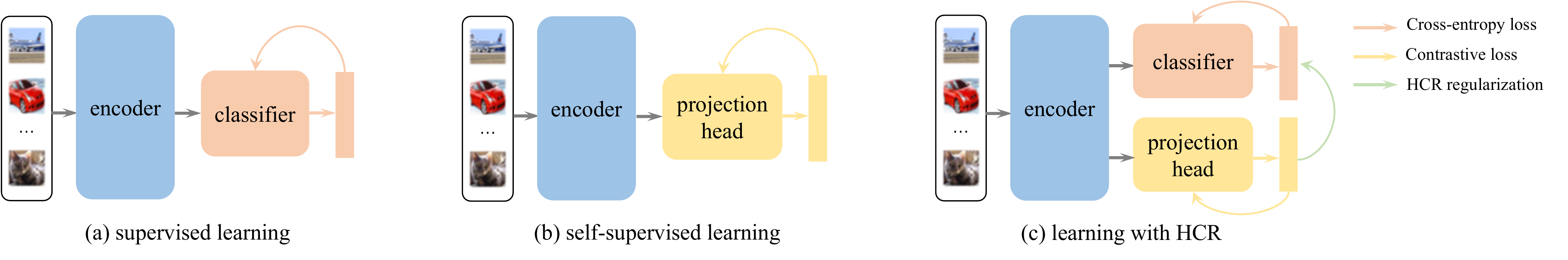}
  \caption{Conceptual illustration of different learning paradigms. We suppose supervised learning and self-supervised learning learn proper representations through different pretext tasks relying on label-dependent and feature-dependent information respectively. HCR takes supervised learning as the primary task and forces self-supervised learning to assist it from another perspective.}
  \label{fig:conceptual_illustration}
\end{figure*}

\section{Related works}

\subsection{Contrastive learning}
Self-supervised learning designs pretext tasks~\cite{zhang2016colorful,pathak2016context,gidaris2018unsupervised,noroozi2016unsupervised} to produce supervision signals derived from the data itself, while contrastive learning is its subset that aims to group similar samples closer and diverse samples away from each other~\cite{wu2018unsupervised,he2020momentum,henaff2020data,chen2020simple,NEURIPS2020_f3ada80d,chen2020improved,chen2020big,caron2020unsupervised,li2020prototypical,wang2020understanding,zoph2020rethinking,hu2021adco,zbontar2021barlow,tianyuan_directpred}. Inspired by the important technique $\ell_2$ normalization on metric learning~\cite{schroff2015facenet,liu2017sphereface,wang2017normface}, \cite{wu2018unsupervised} takes the class-wise supervision to the extreme of instance-wise supervision and tries to maximally scatter the features of samples over the unit hypersphere. Most of the subsequent works~\cite{he2020momentum,henaff2020data,chen2020simple,NEURIPS2020_f3ada80d,chen2020improved,chen2020big,caron2020unsupervised,li2020prototypical,hu2021adco} on contrastive learning employ $\ell_2$ normalization as a standard setting, while \cite{wang2020understanding} highlights $\ell_2$ normalization helps contrastive learning optimize uniformity of the induced distribution of the features on the hypersphere together with the alignment of features from positive pairs. Representation learning benefits from the desirable traits of placing features on the unit hypersphere that improving training stability and separable ability. 

Extending contrastive learning to semi-supervised learning or weakly-supervised learning is straightforward. SupCon~\cite{khosla2020supervised} proposes class-wise contrastive loss under fully-supervised setting and inspires researchers to focus on the power of contrastive learning in supervised scenarios. Self-Tuning~\cite{wang2021self} explores group contrastive learning and tackles confirmation bias and model shift issues in an efficient one-stage framework towards data-efficient transfer learning and semi-supervised learning. CoMatch~\cite{li2020comatch} unifies contrastive learning, consistency regularization, entropy minimization and graph-based SSL to mitigate confirmation bias in pseudo-label-based semi-supervised learning. PSC~\cite{wang2021contrastive} proposes a hybrid network that jointly performs both self-supervised learning and prototypical supervised contrastive learning in a cumulative learning manner. BalFeat~\cite{kang2020exploring} combines strengths of supervised methods and contrastive methods to learn representations that are both discriminative and balanced. MoPro~\cite{li2020mopro} simultaneously optimizes classical supervised loss and prototypical contrastive loss using momentum prototypes and tries to achieve robust weakly-supervised learning. Co-learning~\cite{tan2021co} rethinks that the co-training-based noisy label learning methods provide limited information gain since the differences between two networks of the same architecture mainly come from random initialization. Thus, this method explores intrinsic similarity and structural similarity to combat noisy labels. 

These methods~\cite{wang2021self,wang2021contrastive,kang2020exploring,li2020comatch, tan2021co, cheng2021demystifying} have similar architectures that concurrently optimize typical contrastive learning and supervised learning with certain techniques. We can see this manner from a different perspective: both label-dependent supervised learning and feature-dependent contrastive learning are pretext tasks that aim to learn proper representations. Previous works~\cite{doersch2017multi,bachman2019learning} combine different pretext tasks to boost self-supervised learning performance and find there exist relationships between pretext tasks in which regularization is in need. Thus, our method regularizes the implicit connections between supervised learning and self-supervised learning in hypersphere space as shown in Figure~\ref{fig:conceptual_illustration}. HCR builds a bridge between classical supervised learning and pretext tasks in self-supervised learning, and the regularization is plug-and-play to apply in these joint-learning methods.

\subsection{Learning on the hypersphere} 
There are quite a number of methods that learn representations on the hypersphere~\cite{liu2016large,liu2017deep,liu2018learning,wang2018cosface,davidson2018hyperspherical,deng2019arcface,park2019sphere,lin2020regularizing,chen2020angular,liu2021learning,liu2021orthogonal,xu2019larger,jing2021balanced} and show that the key semantics in neural networks is angular information instead of magnitude. MHE~\cite{liu2018learning} draws inspiration from the Thomson problem to regularize networks with a minimum hyperspherical energy objective for improving the generalization ability of networks. CoMHE~\cite{lin2020regularizing} shows that naively minimizing hyperspherical energy suffers from difficulties due to highly nonlinear and non-optimization, and proposes projecting neurons to suitable subspaces where hyperspherical energy can get minimized efficiently. Moreover, Johnson-Lindenstrauss lemma~\cite{dasgupta2003elementary} establishes a guarantee for CoMHE's projections. SphereGAN~\cite{park2019sphere} remaps Euclidean feature spaces into the hypersphere by geometric transformation and calculates geometric moments for minimizing the multiple Wasserstein distances of probability measures on the hypersphere. Our work reprojects the Euclidean feature space of the classifier into a hypersphere and explores its connection with the projection head's hypersphere.

\section{Methods} 

\subsection{Preliminaries}
HCR focuses on regularizing the manner that is jointly training supervised learning and self-supervised learning and tries to find their relationships. Suppose $\mathcal{X} \subset \mathbb{R}^n$ is the $n$-dimensional Euclidean image space, and $\mathcal{Y} = \{0, 1\}^c$ is the ground-truth label space with $c$ classes in an one-hot manner if the label exists. The usual framework consists of a classifier $g: \mathbb{R}^{D_f} \rightarrow \mathbb{R}^{D_g}$ and a projection head $h: \mathbb{R}^{D_f} \rightarrow \mathbb{S}^{D_h-1}$ with their shared feature encoder $f: x \in \mathcal{X} \rightarrow \mathbb{R}^{D_f}$, where $D_f$, $D_g$, $D_h$ denote the dimension of output Euclidean spaces from $f$, $g$, $h$ respectively. HCR imposes constraints in hyperspherical spaces so that the classifier $g: \mathbb{R}^{D_f} \rightarrow \mathbb{S}^{D_g-1}$ outputs a $(D_g-1)$-dimensional hypersphere $\mathbb{S}^{D_g-1}$ by mapping the original outputs to $\ell_2$ normalized feature vectors of dimension $D_g$.

\subsection{Hyperspherical consistency regularization}
\begin{figure}[ht]
  \centering
  \includegraphics[width=0.48\textwidth]{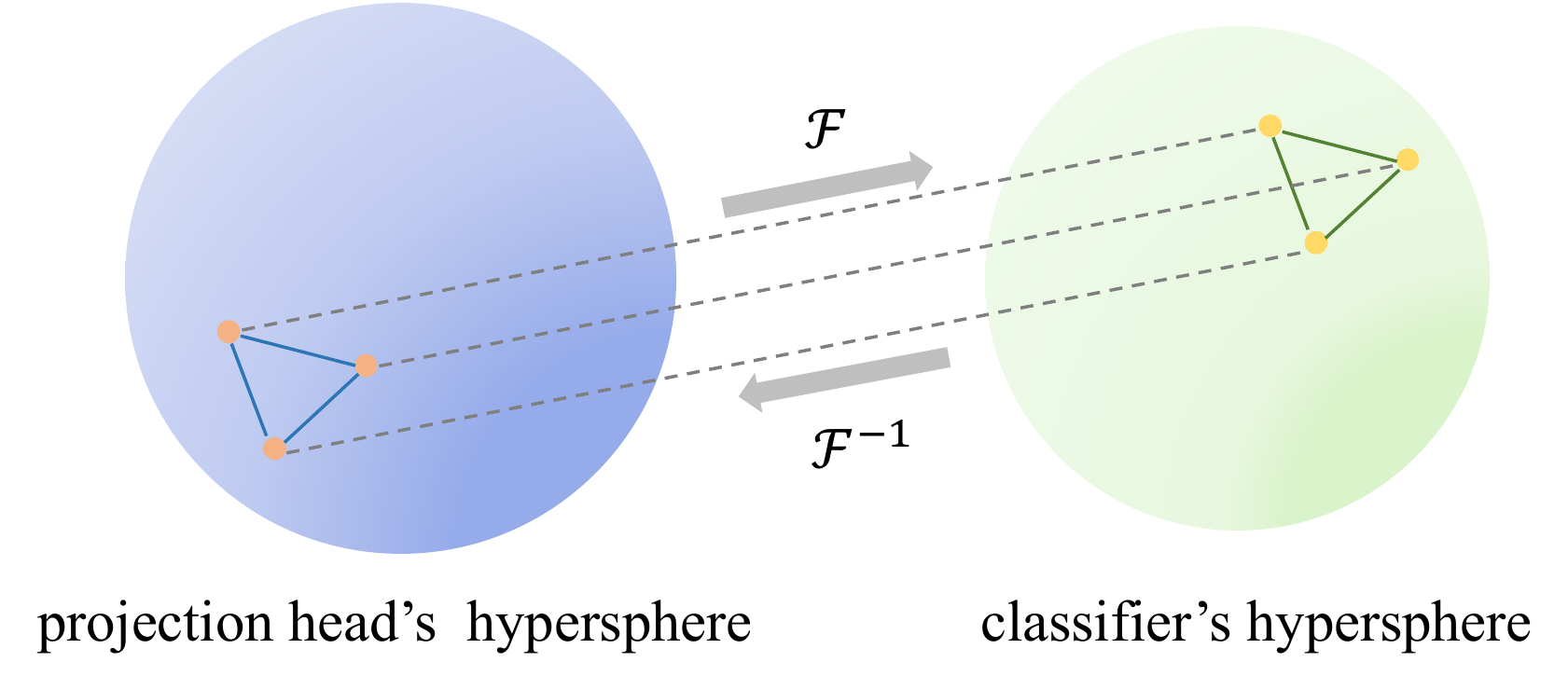}
  \caption{Preserve the geometry structure of data points lies on hyperspheres by their pairwise distances.}
  \label{fig:preserve_structure}
\end{figure}

As the classifier $g(\cdot)$ and the projection head $h(\cdot)$ perform different tasks according to the same features from the feature encoder $f(\cdot)$, HCR assumes there exists a distance-preserving mapping $\mathcal{F}: \mathbb{R}^{D_h} \rightarrow \mathbb{R}^{D_g}$ and its inverse mapping $\mathcal{F}^{-1}: \mathbb{R}^{D_g} \rightarrow \mathbb{R}^{D_h}$ that establish the connections of points on the hyperspheres. We argue that the relationship of the hyperspheres from different tasks can be characterized by the geometric property. Here, we consider the pairwise distance as the key geometry property, and force points on the classifier's hypersphere to have a similar structure as the projection head's, as shown in Figure~\ref{fig:preserve_structure}. 

We first define the pairwise distances on the respective hyperspheres as:
\begin{equation}
  \begin{aligned}
    d_g(x_i, x_j) = \|g \circ	f(x_i) - g \circ f(x_j)\|,\\
    d_h(x_i, x_j) = \|h \circ	f(x_i) - h \circ f(x_j)\|,
  \end{aligned}
\end{equation}
where $\|\cdot\|$ denotes Euclidean distance. While $x_i, x_j \in \mathcal{X}$ and $i \neq j$, we use $d_g$ and $d_h$ to represent the sets of pairwise distances on $\mathcal{X}$ for notation simplicity. To measure the pairwise distances on hyperspheres with different dimensions, we define similarity metrics $p(d_g)$ and $q(d_h)$ that are considered to be normal distributions multiplied by constant terms (see Sec \ref{distribution-of-distance}):

\begin{equation}
  \begin{aligned}
    p(d_g) = C_g \frac{1}{\sigma_g\sqrt{2 \pi}} \exp\left[{-\frac{1}{2}\frac{(d_g - \mu_g)^2}{\sigma_g^2}}\right], \\
    q(d_h) = C_h \frac{1}{\sigma_h\sqrt{2 \pi}} \exp\left[{-\frac{1}{2}\frac{(d_h - \mu_h)^2}{\sigma_h^2}}\right],
  \end{aligned}
\end{equation}
where $C_g, C_h$ are constants that forces the similarity metric to be in $[0, 1]$. $\sigma_g, \mu_g, \sigma_h, \mu_h$ can be chosen according to the situations. For the convenience of optimization, we empirically assumes $p(d_g), q(d_h) \sim N(0, \frac{1}{2})$ in all experiments except especially mentioned.

As contrastive learning tries to push away samples from different classes and pull together samples from the same classes, the respective similarity metrics are supposed to be approaching either zero or one. Thus, we define the objective of HCR that minimizing the binary cross entropy (BCE) between $p(d_g)$ and $q(d_h)$:

\begin{equation}
  \begin{aligned}
    & \mathrm{HCR}(p(d_g), q(d_h)) = \mathrm{BCE}(p(d_g) \; || \; q(d_h)) \\ &= - p(d_g) \log q(d_h) - (1 - p(d_g)) \log (1 - q(d_h)),
  \end{aligned}
  \label{hcr}
\end{equation}
Through minimizing Equation~\ref{hcr}, the mutual information $I(g \circ f(x), h \circ f(x))$ between logits $g \circ f(x)$ and feature projections $h \circ f(x)$ is implicitly maximized (see Sec \ref{mutual-information}).


\subsection{HCR as a regularization for learning}
Now that we have introduced the formulation of HCR, here we propose HCR as a regularization for semi-supervised or weakly-supervised learning. While HCR imposes the consistency between the classifier and the projection head on the hyperspherical latent spaces, it is suitable for the jointly learning manner that performs supervised learning and self-supervised learning simultaneously. In such a setting, the entire objective function can be represented as:

\begin{equation}
  \begin{aligned}
    \mathcal{L} = \sum_{(x, y) \in \mathcal{X} \times \mathcal{Y}} \mathcal{L}_s(x, y) + \sum_{x \in \mathcal{X}} \mathcal{L}_u(x) \\ + \; \mathrm{HCR}(p(d_g), q(d_h)),
    \label{semi-loss}
  \end{aligned}
\end{equation}
where $\mathcal{L}_s$ denotes the supervised loss for the labeled data, and $\mathcal{L}_u$ denotes the contrastive loss (i.e., the commonly-used InfoNCE~\cite{oord2018representation}) for the unlabeled data. HCR regularizes supervised learning and explores its connections with contrastive learning so that both $\mathcal{L}_s$ and $\mathcal{L}_u$ are needed.



\subsection{Relations to supervised contrastive learning}
Similar to SupCon~\cite{khosla2020supervised}, HCR leverages contrastive learning to benefit vanilla supervised learning. While SupCon explicitly pushes apart clusters of samples from different classes and pulls together clusters of samples belonging to the same class in a self-supervised contrastive manner, HCR implicitly makes agreements on latent hyperspherical spaces between contrastive learning and supervised learning. Moreover, SupCon directly imposes contrastive loss to standard cross entropy, which requires a full exploration of label information so that it is limited in supervised learning fashion. HCR forces supervised learning to imitate contrastive learning in latent spaces without extra label information so that it can conveniently fit into semi-supervised learning and weakly-supervised learning frameworks.

\section{Theoretical Insights}

This section is inspired by rigorous theoretical results from \cite{hammersley1950distribution, lord1954distribution, alagar1976distribution,kuijlaars1998asymptotics,johnson1984extensions,dasgupta2003elementary,baraniuk2008simple,candes2006near,durrant2012random,kraskov2004estimating,tschannen2019mutual,cover2006elements} and provides theoretical and intuitive perspectives about HCR.

\subsection{Distribution of distance on a hypersphere}
\label{distribution-of-distance}
\paragraph{Theorem 1.} (Asymptotic form of the distribution of Euclidean distance in a hypersphere for large dimensions). \textit{Given the $D$-dimensional hypersphere $\mathbb{S}^D(a)$ with radius $a$, $x_i$ and $x_j (\forall i \neq j)$ are any two points chosen at random in $\mathbb{S}^n(a)$ whose Euclidean distance is denoted by $r (0 \leq r \leq 2a)$. Then, the asymptotic distribution of $r$ is $N(\sqrt{2}a, \frac{a^2}{2D})$ as $D \rightarrow \infty$.}

Theorem 1, which has been heavily studied in \cite{hammersley1950distribution, lord1954distribution, alagar1976distribution}, tells us the distribution of Euclidean distance in a hypersphere obeys normal distribution as the dimension of the hypersphere becomes large. Though HCR only considers the case that points on the hypersphere surface and ignores the points inside, it still agrees with this theorem. HCR models the pairwise distance distribution as normal distribution and tries to utilize pairwise distances as a key property for preserving the geometric structure of the projection head's outputs. Thus, HCR builds a bridge between feature-dependent information and label-dependent information.

\subsection{Connections between hyperspheres}
\label{jll-lemma}
\paragraph{Theorem 2.} (Johnson-Lindenstrauss lemma). \textit{Let $\epsilon \in (0, 1)$. Let $N, D_g \in \mathbb{N}$ such that $D_g \leq C \epsilon^{-2} \log N$, for a large enough absolute constant C. Let $H \subseteq \mathbb{R}^{D_h}$ be a set of $N$ points. There exists a linear mapping $\mathcal{F}: \mathbb{R}^{D_h} \rightarrow \mathbb{R}^{D_g}$, such that for all $h_i, h_j \in H$:}
\begin{equation}
  (1 - \epsilon)||h_i - h_j||^2 \leq ||\mathcal{F}\circ h_i - \mathcal{F}\circ h_j||^2 \leq (1 + \epsilon)||h_i - h_j||^2.
\end{equation}
The famous Johnson-Lindenstrauss lemma has found numerous applications that includes searching for graph embedding, manifold learning and dimension reduction. Here, this lemma guarantees the projection of two points from high-dimensional space to low-dimensional space preserves their Euclidean distance with high probability. Though the dimensions of feature projections $h \circ f(x)$ and logits $g \circ f(x)$ are usually not the same, HCR preserves their relative distances under this theorem.

\label{mutual-information}
\paragraph{Theorem 3.} (Mutual information's invariance property to reparametrization of the marginal variables). If $H' = \mathcal{F}(H)$ and $G' = \mathcal{T}(G)$ are homeomorphisms (i.e. smooth uniquely invertible maps), then the mutual information $I(H, G) = I(H', G')$.

This theorem~\cite{kraskov2004estimating,tschannen2019mutual,cover2006elements} reveals the possible connections beween $g \circ f(x)$ and $h \circ f(x)$. We discuss that Equation~\ref{hcr} is preserving pairwise distances between the projection head's and the classifier's hyperspherical spaces. For those limited data points, we consider the mapping $\mathcal{F}$ from the projection head $h$ to the classifier $g$ is approaching bijective through preserving pairwise distances. Since when $\mathcal{F}$ is invertible, the mutual information is:
\begin{equation}
  \begin{aligned}
    I(h \circ f(x), g \circ f(x)) &= I(h \circ f(x), \mathcal{F}\circ (h \circ f(x))) \\ 
    &= I(h \circ f(x), h \circ f(x))
  \end{aligned}
\end{equation}
so that it is maximized. Thus, HCR preserves the distributions of pairwise distances that implicitly maximizes the mutual information $I(h \circ f(x), g \circ f(x))$.

\section{Experiments} 
To validate the effectiveness of our proposed HCR, we conduct experiments on various tasks, such as semi-supervised learning, fine-grained classification, and noisy label learning, among which the latter two tasks belong to weakly-supervised learning.

\subsection{Baselines}

We take the recent typical works Self-Tuning~\cite{wang2021self} and Co-learning~\cite{tan2021co} as our baselines on account of their jointly learning manner, that is, both of them build the network architecture using a shared feature encoder with a classifier and a projector head and train two heads simultaneously though in different ways. 

\paragraph{Self-Tuning} unifies the exploration of labeled data and unlabeled data and the transfer of a pretrained model in a pseudo group contrast (PGC) mechanism. The vanilla contrastive learning maximizes the similarity between query $q$ with its corresponding positive key $k_0$ (a different view of the same data sample):

\begin{equation}
  \mathcal{L}_{\mathrm{CL}} = -\log \frac{\exp(q \cdot k_0 / \tau)}{\exp(q \cdot k_0 / \tau) + \sum_{d=1}^D \exp(q \cdot k_d / \tau)},
  \label{eq-cl}
\end{equation}
where $\tau$ is a hyperparameter for temperature scaling.
Self-Tuning modifies the contrastive mechanism by introducing a group of positive keys from samples with the same pseudo label to contrast with other samples as follows:

\begin{equation}
  \mathcal{L}_{\mathrm{PGC}} = - \frac{1}{D+1} \sum_{d=0}^D \log \frac{\exp(q \cdot k_d^{\hat{y}} / \tau)}{\exp(q\cdot k_0^{\hat{y}/\tau}) + \mathrm{Neg}},
  \label{eq-pgc}
\end{equation}
where $\mathrm{Neg} = \sum_{c=1}^{\{1,2,...,C\}\backslash\hat{y}} \sum_{j=1}^D \exp(q \cdot k_j^c / \tau)$, $D$ is the number of classes, $\hat{y}$ denotes the pseudo label. Equation~\ref{eq-cl} and Equation~\ref{eq-pgc} are corresponding to the unlabeled loss $\mathcal{L}_u$ in Equation~\ref{semi-loss}.

\begin{figure}[ht]
  \centering
  \includegraphics[width=0.46\textwidth]{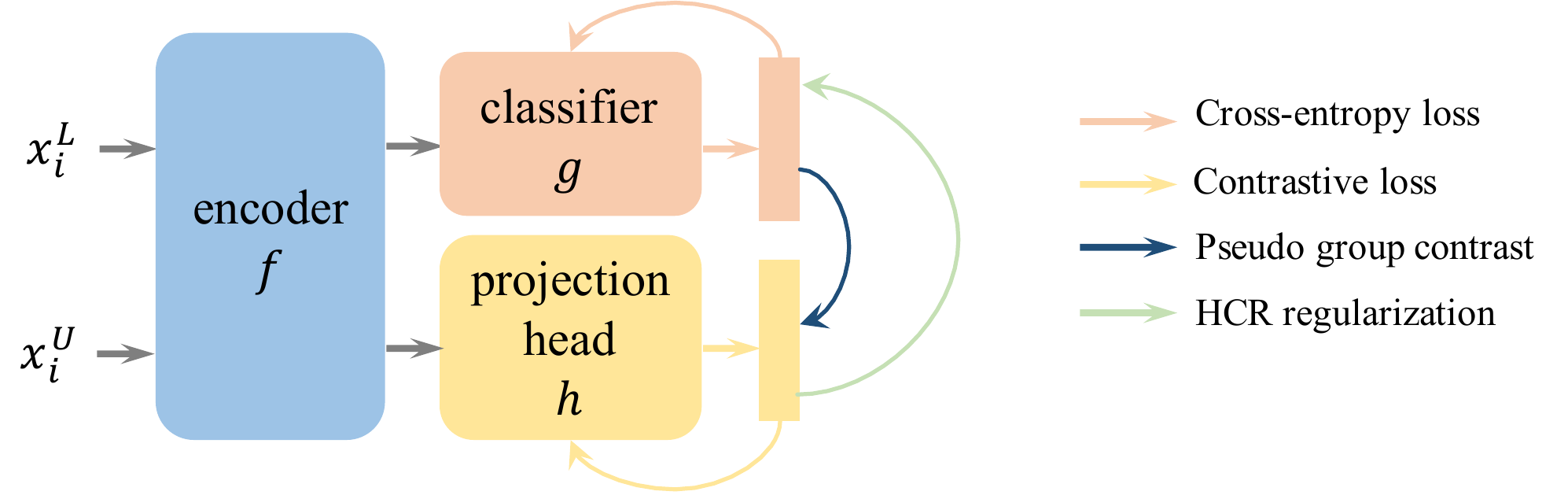}
  \caption{The illustration of Self-Tuning with HCR. $x_i^L$ denotes data samples from the labeled dataset, and $x_i^U$ denotes data samples from the unlableed dataset. PGC uses pseudo label information from the classifier to guide the projection head, while HCR reversely uses projections to correct the classifier.}
  \label{fig:self-tuning-with-hcr}
\end{figure}

Self-Tuning has bridged supervised learning and self-supervised learning by guiding the contrastive mechanism of the projection head through pseudo labels produced by the classifier. However, this scheme can only help the feature encoder obtain decent representations while the bias of the classifier remains unavoidable. Thus, we expect the projection head to benefit the classifier as well through HCR as shown in Figure~\ref{fig:self-tuning-with-hcr}.

\paragraph{Co-learning} is a recent work that challenges the co-training scheme in noisy label learning. It provides perspectives from both supervised learning and self-supervised learning through the above mentioned jointly learning manner. As the classifier is extremely unreliable in noisy learning settings, Co-learning also proposes a structural similarity that imposes structure-preserving constraints similar to HCR. Co-learning directly assumes the pairwise distance follows a normal distribution and minimizes the Kullback–Leibler (KL) divergence of the distributions between the projection head and the classifier. 

\begin{figure*}[ht]
  \centering
  \includegraphics[width=0.99\textwidth]{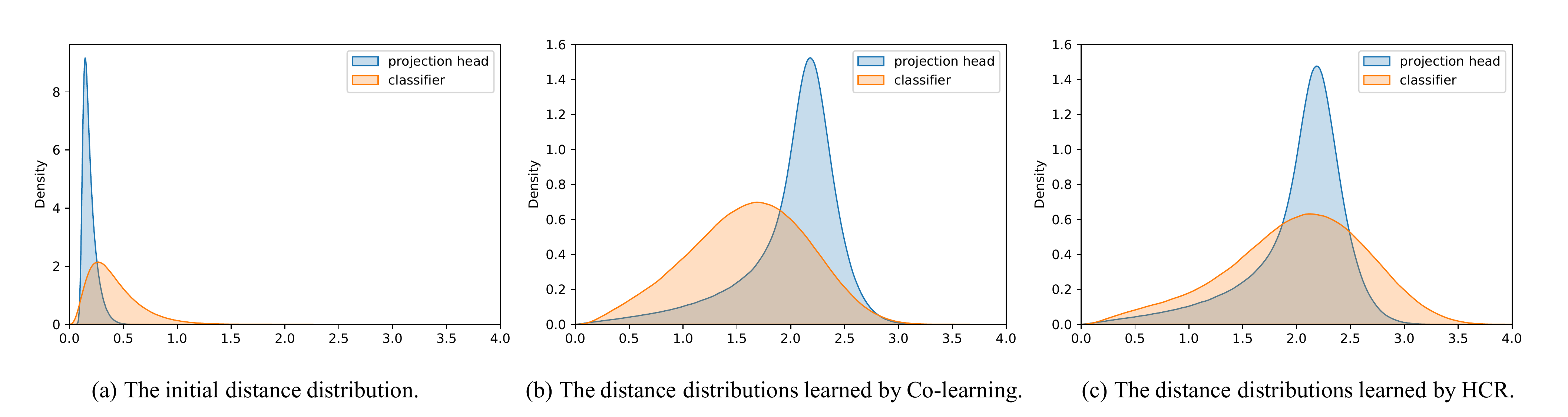}
  \caption{The hyperspherical distance distributions of the CIFAR-10 dataset with 80\% symmetric noise.}
  \label{fig:comparison-distance-distribution}
\end{figure*}

However, Co-learning cannot theoretically guarantees that the structure-preserving constraints are well performed. As shown in Figure~\ref{fig:comparison-distance-distribution}, we train Co-learning and HCR on the CIFAR-10 dataset with 80\% symmetric noise for a hundred epochs, and present the hyperspherical distance distributions. The initial hyperspherical pairwise distances of the projection head are roughly in $[0, 0.5]$. After training, they become large, which suggests the conclusion of \cite{wang2020understanding} is correct, i.e., contrastive learning is trying to make features uniformly distributed on the hypersphere rather than concentrating in a local area. Moreover, the distance distribution learned by HCR preserves the structure better than Co-learning's because the classifier's distance distribution is balanced referring to the projection head's.

\subsection{Semi-supervised learning}

Following the same experimental setting as \cite{wang2021self}, we compare Self-Tuning with HCR against three classical semi-supervised learning methods: Pseudo-Labeling~\cite{lee2013pseudo}, $\Pi$-model~\cite{laine2016temporal}, and Mean Teacher~\cite{tarvainen2017mean}, as well as three recent methods UDA~\cite{xie2020unsupervised}, FixMatch~\cite{sohn2020fixmatch}, SimCLRv2~\cite{chen2020big}, and Self-Tuning itself~\cite{wang2021self}. Experiments are conducted on three mainstream visual datasets: Stanford Cars~\cite{krause20133d}, FGVC Aircraft~\cite{maji2013fine}, CUB-200-2011~\cite{wah2011caltech}, and CIFAR-100~\cite{krizhevsky2009learning}. Stanford Cars contains 16185 images of 196 classes of cars, and the pixel resolution is $360 \times 240$. FGVC Aircraft consists of 10000 images of 100 different aircraft model variants. The image resolution is about 1-2M pixels, but its width and height are not fixed. CUB-200-2011 is a dataset with totally 6033 images of 200 bird species, and each image has about less than 250 thousand pixels. CIFAR-100 is a classical visual dataset with 100 classes and 600 images per class, and the image resolution is $32 \times 32$.

For a fair comparison, all these methods implement a ResNet-50 model and initialize it from ImageNet-pretrained weights. Besides, we remove the last layer of the pretrained model and add the projection head $h$ and classifier $g$ with randomly initialized weights. The default temperature $\tau$ is 0.07, and the learning rate is 0.001. The optimizer follows the original Self-Tuning, which is SGD with a momentum of 0.9. Experiments are repeated three times with different random seeds, and we report the average test accuracy of three trials for each experiment. When we reproduce Self-Tuning, we unexpectedly find the results are better than its paper reported, thus we honestly \textit{report the reproduced results rather than using results from its paper}.

As reported in Table \ref{tab:semi-sup-stanford}, HCR significantly improves the performance of Self-Tuning by an average of 2.30\% in different label proportions. Moreover, HCR obtains averagely 3.77\% improvements under the condition of 15\% label proportion, which indicates the effectiveness of HCR with extremely few labels.

\begin{table}[ht]\footnotesize
\centering
\setlength{\tabcolsep}{2.5mm}{
  \caption{Classification accuracy (\%) $\uparrow$ of semi-supervised learning methods on Stanford Cars dataset (ResNet-50 pretrained).}
  \label{tab:semi-sup-stanford}
  \begin{tabular}{cccc}
    \toprule
    \multirow{2}{*}{Method} & \multicolumn{3}{c}{Label Proportion} \\ 
                            & 15\%           & 30\%           & 50\%       \\ \midrule
    Pseudo-Labeling         & 40.93$\pm$0.23 & 67.02$\pm$0.19 & 78.71$\pm$0.30 \\ 
    $\Pi$-model             & 45.19$\pm$0.21 & 57.29$\pm$0.26 & 64.18$\pm$0.29 \\
    Mean Teacher            & 54.28$\pm$0.14 & 66.02$\pm$0.21 & 74.24$\pm$0.23 \\ 
    UDA                     & 39.90$\pm$0.43 & 64.16$\pm$0.40 & 71.86$\pm$0.56 \\ 
    FixMatch                & 49.86$\pm$0.27 & 77.54$\pm$0.29 & 84.78$\pm$0.33 \\ 
    SimCLRv2                & 45.74$\pm$0.16 & 61.70$\pm$0.18 & 77.49$\pm$0.24 \\ 
    \midrule
    Self-Tuning             & 74.99$\pm$0.11 & 85.87$\pm$0.04 & 89.83$\pm$0.01 \\ 
    Self-Tuning+HCR         & \textbf{78.76}$\pm$0.08 & \textbf{87.70}$\pm$0.07 & \textbf{91.14}$\pm$0.06 \\ 
    \bottomrule
  \end{tabular}
}
\end{table}

Table \ref{tab:semi-sup-aircraft} shows results on FGVC Aircraft dataset. As we can see, the observations are consistently the same as those for Stanford Cars dataset, which is, HCR is still able to obtain large gains (averagely 3.03\%) even Self-Tuning has achieved a very high accuracy. Also, the lower the label proportion, the larger the improvements brought by HCR.

\begin{table}[ht]\footnotesize
\centering
\setlength{\tabcolsep}{2.5mm}{
  \caption{Classification accuracy (\%) $\uparrow$ of semi-supervised learning methods on FGVC Aircraft dataset (ResNet-50 pretrained).}
  \label{tab:semi-sup-aircraft}
  \begin{tabular}{cccc}
    \toprule
    \multirow{2}{*}{Method} & \multicolumn{3}{c}{Label Proportion} \\ 
                            & 15\%           & 30\%           & 50\%       \\ \midrule
    Pseudo-Labeling         & 46.83$\pm$0.30 & 62.77$\pm$0.31 & 73.21$\pm$0.39 \\ 
    $\Pi$-model             & 37.72$\pm$0.25 & 58.49$\pm$0.26 & 65.63$\pm$0.36 \\
    Mean Teacher            & 51.59$\pm$0.23 & 71.62$\pm$0.29 & 80.31$\pm$0.32 \\ 
    UDA                     & 43.96$\pm$0.45 & 64.17$\pm$0.49 & 67.42$\pm$0.53 \\ 
    FixMatch                & 55.53$\pm$0.26 & 71.35$\pm$0.35 & 78.34$\pm$0.43 \\ 
    SimCLRv2                & 40.78$\pm$0.21 & 59.03$\pm$0.29 & 68.54$\pm$0.30 \\ 
    \midrule
    Self-Tuning             & 66.68$\pm$0.17 & 79.94$\pm$0.09 & 84.35$\pm$0.08 \\ 
    Self-Tuning+HCR         & \textbf{70.54}$\pm$0.02 & \textbf{82.64}$\pm$0.04 & \textbf{86.89}$\pm$0.15 \\ 
    \bottomrule
  \end{tabular}
}
\end{table}

We report results on CUB-200-2011 dataset in Table \ref{tab:semi-sup-cub200}. It can be seen that applying HCR yields better performance than the original Self-Tuning. 

\begin{table}[ht]\footnotesize
  \centering
  \caption{Classification accuracy (\%) $\uparrow$ of semi-supervised learning methods on CUB-200-2011 dataset (ResNet-50 pretrained).}
  \setlength{\tabcolsep}{2.5mm}{
    \begin{tabular}{cccc}
      \toprule
      \multirow{2}{*}{Method} & \multicolumn{3}{c}{Label Proportion} \\ 
      & 15\%           & 30\%           & 50\%       \\ \midrule
      Pseudo-Labeling         & 45.33$\pm$0.23 & 56.20$\pm$0.29 & 64.07$\pm$0.32 \\ 
      $\Pi$-model             & 45.20$\pm$0.25 & 58.49$\pm$0.26 & 65.63$\pm$0.36 \\
      Mean Teacher            & 53.26$\pm$0.19 & 66.66$\pm$0.20 & 74.37$\pm$0.30 \\ 
      UDA                     & 46.90$\pm$0.31 & 61.16$\pm$0.35 & 71.86$\pm$0.43 \\ 
      FixMatch                & 44.06$\pm$0.23 & 63.54$\pm$0.18 & 75.96$\pm$0.29 \\ 
      SimCLRv2                & 45.74$\pm$0.15 & 62.70$\pm$0.24 & 71.01$\pm$0.34 \\ 
      \midrule
      Self-Tuning             & 64.79$\pm$0.06 & 74.31$\pm$0.07 & 78.45$\pm$0.31 \\ 
      Self-Tuning+HCR         & \textbf{66.42}$\pm$0.24 & \textbf{75.06}$\pm$0.13 & \textbf{79.48}$\pm$0.16 \\ 
      \bottomrule
    \end{tabular}
  }
  \label{tab:semi-sup-cub200}
  \end{table}

\begin{figure}[ht]
  \centering
  \includegraphics[width=0.48\textwidth]{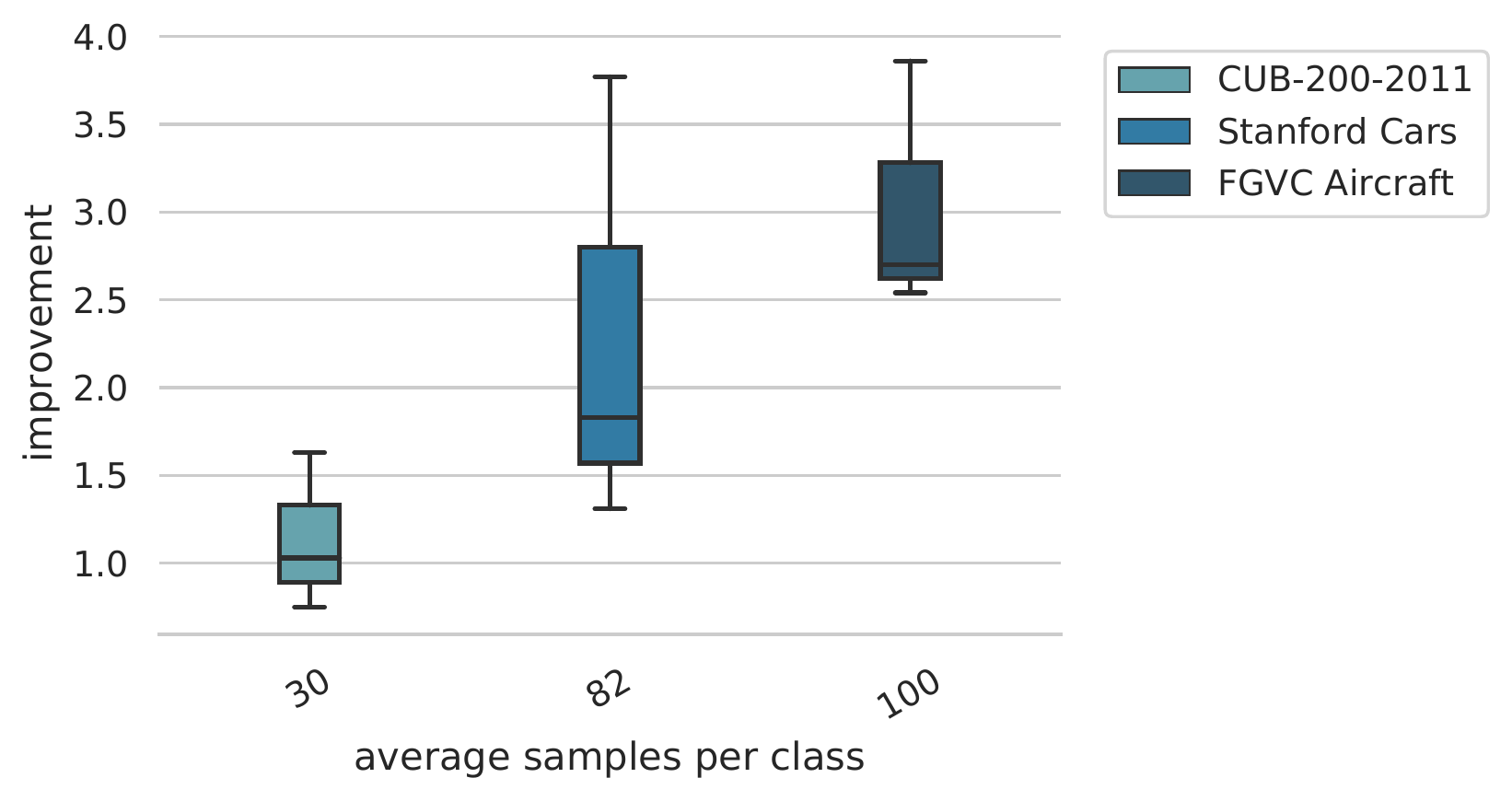}
  \caption{The improvements brought by HCR are proportional to the average number of samples per class.}
  \label{fig:average-sample-per-class}
\end{figure}

Note that the improvement on CUB-200-2011 dataset is slightly less than the former two datasets. The reason is that the average number of samples per class of CUB-200-2011 is much less than Stanford Cars and FGVC Aircraft datasets. It's difficult for HCR to capture the structure of data while those data are sparsely distributed on the hypersphere. We analyze the relationship between average samples per class and the improvements as shown in Figure~\ref{fig:average-sample-per-class}.

Except for the visual datasets evaluated in Self-Tuning, we perform experiments on the standard semi-supervised learning benchmark CIFAR-100 dataset. The results are shown in Table \ref{tab:semi-sup-cifar100-1} and Table \ref{tab:semi-sup-cifar100-2}. Self-Tuning implements EfficientNet-B2 model~\cite{tan2019efficientnet} as the pretrained weights of WRN-28-8~\cite{Zagoruyko2016WRN} are not available. While FixMatch obtains a higher error rate with EfficientNet-B2 than WRN-28-8, Self-Tuning outperforms those methods on WRN-28-8. HCR here further increases its leading.

\begin{table}[ht]\footnotesize
\centering
\caption{Error rates (\%) $\downarrow$ of semi-supervised learning methods on CIFAR-100 dataset with 2500 labes, and 10000 labels.}
\label{tab:semi-sup-cifar100-1}
\setlength{\tabcolsep}{4mm}{
  \begin{tabular}{lcll}
    \toprule
    \multicolumn{1}{c}{\multirow{2}{*}{Method}} & \multirow{2}{*}{Network}                                                                 & \multirow{2}{*}{2.5K} & \multicolumn{1}{c}{\multirow{2}{*}{10k}} \\
    \multicolumn{1}{c}{}    
    & & & \multicolumn{1}{c}{} \\
    \midrule
    Pseudo-Labeling &  & 57.38 & 36.21     \\
    $\Pi$-Model   & \multirow{7}{*}{\begin{tabular}[c]{@{}c@{}}WRN-28-8\\ \#Para: 11.76M\end{tabular}}       & 57.25                 & \multicolumn{1}{c}{37.88}                \\
    Mean Teacher    &  & 53.91 & 35.83 \\
    MixMatch        &  & 39.94 & 28.31 \\
    UDA             &  & 33.13 & 24.50 \\
    ReMixMatch      &  & 27.43 & 23.03 \\
    FixMatch        &  & 28.64 & 23.18 \\
    \midrule
    FixMatch        & \multirow{5}{*}{\begin{tabular}[c]{@{}c@{}}EfficientNet-B2\\ \#Para: 9.43M\end{tabular}} & 29.99 & 21.69 \\

    Fine-Tuning     &  & 31.69 & 21.74 \\
    Co-Tuning       &  & 30.94 & 22.22 \\
    Self-Tuning     &  & 24.16 & 17.57 \\
    Self-Tuning+HCR &  & \textbf{23.93} & \textbf{16.24}  \\      
    \bottomrule                           
    \end{tabular}
}
\end{table}
  
Besides, under extremely few labels condition, HCR strongly outperforms other methods through promoting Self-Tuning by 4.46\% which is definitely a large margin. We believe our proposed HCR can play a significant role in bridging feature-dependent and label-dependent information, especially in the case of few labels being available.

\begin{table}[ht]\footnotesize
\centering
\caption{Error rates (\%) $\downarrow$ of semi-supervised learning methods on CIFAR-100 dataset with only 400 labels (EfficientNet-B2 pretrained). CT: Co-Tuning, PL: Pseduo Labeling, MT: MeanTeacher, FM: FixMatch.}
\label{tab:semi-sup-cifar100-2}
\setlength{\tabcolsep}{2mm}{
\begin{tabular}{l|l|l|l}
\toprule
Fine-Tuning & $L^2$-SP       & DELTA        & BSS      \\
\midrule
60.79       & 59.21          & 58.23        & 58.49    \\
\midrule
Co-Tuning   & Pseudo Labeling &  $\Pi$-model & Mean Teacher \\
\midrule
57.58       & 59.21          & 60.50        & 60.68  \\
\midrule
FixMatch    & UDA            & SimCLRv2     & CT+PL  \\
\midrule
57.87       & 58.32          & 59.45        & 56.21 \\
\midrule
CT+MT       & CT+FM          & Self-Tuning  & Self-Tuning+HCR \\
\midrule
56.78       & 57.94          & 47.17       & \textbf{42.71}  \\
\bottomrule        
\end{tabular}
}
\label{tab:}
\end{table}

\subsection{Fine-grained classification}

We conduct experiments on fine-grained classification using fully-labeled Stanford Cars, FGVC Aircraft, and CUB-200-2011 datasets. The results in Table \ref{tab:fine-grained} show that HCR performs consistently better than the baseline. FGVC Aircraft still gets the most improvements, benefited from its large average number of samples per class as we have mentioned above. We believe HCR can not only help semi-supervised conditions but also difficult supervised learning.

\begin{table}[ht]\footnotesize
\centering
\caption{Classification accuracy (\%) $\uparrow$ of transfer learning methods on fine-grained datasets.}
\setlength{\tabcolsep}{2.5mm}{
\begin{tabular}{lcccc}
\toprule
  Method & Stanford Cars & Aircraft & CUB200\\
\midrule
  Fine-Tuning       & 87.20$\pm$0.19 & 81.13$\pm$0.21 & 78.01$\pm$0.16 \\
  $\mathrm{L}^2$-SP & 86.58$\pm$0.26 & 80.98$\pm$0.29 & 78.44$\pm$0.17 \\
  DELTA             & 86.32$\pm$0.20 & 80.44$\pm$0.20 & 78.63$\pm$0.18 \\
  BSS               & 87.63$\pm$0.27 & 81.48$\pm$0.18 & 78.85$\pm$0.31 \\
  Co-Tuning         & 89.53$\pm$0.09 & 83.87$\pm$0.09 & 81.24$\pm$0.14 \\
\midrule
  Self-Tuning       & 92.33$\pm$0.10 & 88.96$\pm$0.21 & 81.60$\pm$0.11\\
  Self-Tuning+HCR & \textbf{93.03}$\pm$0.06 & \textbf{90.41}$\pm$0.03 & \textbf{82.63}$\pm$0.19\\
\bottomrule
\end{tabular}}
\label{tab:fine-grained}
\end{table}

\subsection{Noisy label learning}
We follows the same experimental settings as \cite{tan2021co} to compare Co-learning with HCR against other co-training-based noisy label learning methods: Decoupling~\cite{malach2017decoupling}, Co-teaching~\cite{han2018co}, Co-teaching+~\cite{yu2019does}, JoCoR~\cite{wei2020combating}, and Co-learning itself~\cite{tan2021co}. We conduct experiments on CIFAR-100 dataset with three different types of noise, i.e., \textit{symmetric}, \textit{asymmetric}, and \textit{instance-dependent}. The details of these noise types are in Appendix. Among these noisy types, we recognize instance-dependent (or feature-dependent) noise as a more realistic setting because human annotations are prone to different levels of errors for tasks with varying difficulty levels. Following~\cite{tan2021co}, we report the average test accuracy over the last 10 epochs of five trials for each experiment. The base model is ResNet-18.

Tabel \ref{tab:cifar100-sym-noise} shows the results on symmetric noise. As it is the simplest synthetic noise type, we perform experiments on high noise ratios as 50\% and 80\%. While Co-learning has already shown amazing results on high noise ratios, HCR further improves Co-learning by averagely 3.92\% under different noise ratios.

\begin{table}[ht]\footnotesize
\centering
\setlength{\tabcolsep}{2.5mm}{
  \caption{Average test accuracy (\%) on CIFAR-100 with symmetric noise over the last 10 epochs.}
\label{tab:cifar100-sym-noise}
\begin{tabular}{lcccc}
\toprule
  Method & sym-20\% & sym-50\% & sym-80\%\\
\midrule
  Standrad CE       & 57.79$\pm$0.44 & 33.75$\pm$0.46 & 8.64$\pm$0.22 \\
  Decoupling        & 56.18$\pm$0.32 & 31.58$\pm$0.54 & 7.71$\pm$0.23 \\
  Co-teaching       & 64.28$\pm$0.32 & 32.62$\pm$0.51 & 6.65$\pm$0.71 \\
  Co-teaching+      & 55.40$\pm$0.71 & 26.49$\pm$0.45 & 8.57$\pm$1.55 \\
  JoCoR             & 62.29$\pm$0.71 & 30.19$\pm$0.60 & 6.84$\pm$0.92 \\
\midrule
  Co-learning       & 66.58$\pm$0.15 & 55.54$\pm$0.43 & 35.45$\pm$0.79 \\
  Co-learning+HCR   & \textbf{70.27}$\pm$0.32 & \textbf{59.93}$\pm$ 0.25 & \textbf{39.14}$\pm$0.47 \\
\bottomrule
\end{tabular}}
\end{table}

We report the results on asymmetric noise in Table \ref{tab:cifar100-asym-noise}. HCR significantly improves Co-learning by 3.74\% on average. Moreover, HCR presents more stable results than Co-learning as the standard deviations are minor.

\begin{table}[ht]\footnotesize
\centering
\caption{Average test accuracy (\%) on CIFAR-100 with asymmetric noise over the last 10 epochs.}
\setlength{\tabcolsep}{2.5mm}{
\begin{tabular}{lcccc}
\toprule
  Method & asym-20\% & asym-30\% & asym-40\%\\
\midrule
  Standrad CE       & 59.36$\pm$0.36 & 51.06$\pm$0.44 & 42.49$\pm$0.23 \\
  Decoupling        & 57.97$\pm$0.24 & 49.86$\pm$0.54 & 41.51$\pm$0.67 \\
  Co-teaching       & 59.76$\pm$0.53 & 49.53$\pm$0.79 & 40.62$\pm$0.79 \\
  Co-teaching+      & 56.11$\pm$0.60 & 47.12$\pm$0.73 & 38.98$\pm$0.54 \\
  JoCoR             & 58.58$\pm$0.51 & 49.04$\pm$0.91 & 39.72$\pm$0.76 \\
\midrule
  Co-learning       & 65.26$\pm$0.76 & 56.97$\pm$1.22 & 47.62$\pm$0.79 \\
  Co-learning+HCR   & \textbf{68.85}$\pm$0.22 & \textbf{61.94}$\pm$0.17 & \textbf{50.29}$\pm$0.69 \\
\bottomrule
\end{tabular}}
\label{tab:cifar100-asym-noise}
\end{table}

When it comes to instance-dependent noise, HCR assists Co-learning in obtaining 1.52\% gains averagely as shown in Table \ref{tab:cifar100-ins-noise}. As Co-learning has utilized feature-dependent information, HCR's improvements on instance-dependent noise are slightly less than the former two noise types.

\begin{table}[ht]\footnotesize
\centering
\caption{Average test accuracy (\%) on CIFAR-100 with instance-dependent noise over the last 10 epochs.}
\setlength{\tabcolsep}{2.5mm}{
\begin{tabular}{lcccc}
\toprule
  Method & ins-20\% & ins-30\% & ins-40\%\\
\midrule
  Standrad CE       & 55.45$\pm$0.54 & 48.77$\pm$0.47 & 41.30$\pm$0.27 \\
  Decoupling        & 52.20$\pm$0.48 & 45.32$\pm$0.83 & 36.33$\pm$0.47 \\
  Co-teaching       & 55.16$\pm$0.61 & 45.24$\pm$0.37 & 34.64$\pm$1.00 \\
  Co-teaching+      & 50.37$\pm$0.85 & 40.73$\pm$0.58 & 32.15$\pm$0.80 \\
  JoCoR             & 54.21$\pm$0.34 & 45.03$\pm$0.52 & 34.08$\pm$1.05 \\
\midrule
  Co-learning       & 69.42$\pm$0.42 & 65.45$\pm$0.86 & 60.40$\pm$1.37 \\
  Co-learning+HCR   & \textbf{70.03}$\pm$0.31 & \textbf{66.89}$\pm$0.41 & \textbf{62.91}$\pm$0.84 \\
\bottomrule
\end{tabular}}
\label{tab:cifar100-ins-noise}
\end{table}

\begin{figure}[ht]
  \centering
  \includegraphics[width=0.45\textwidth]{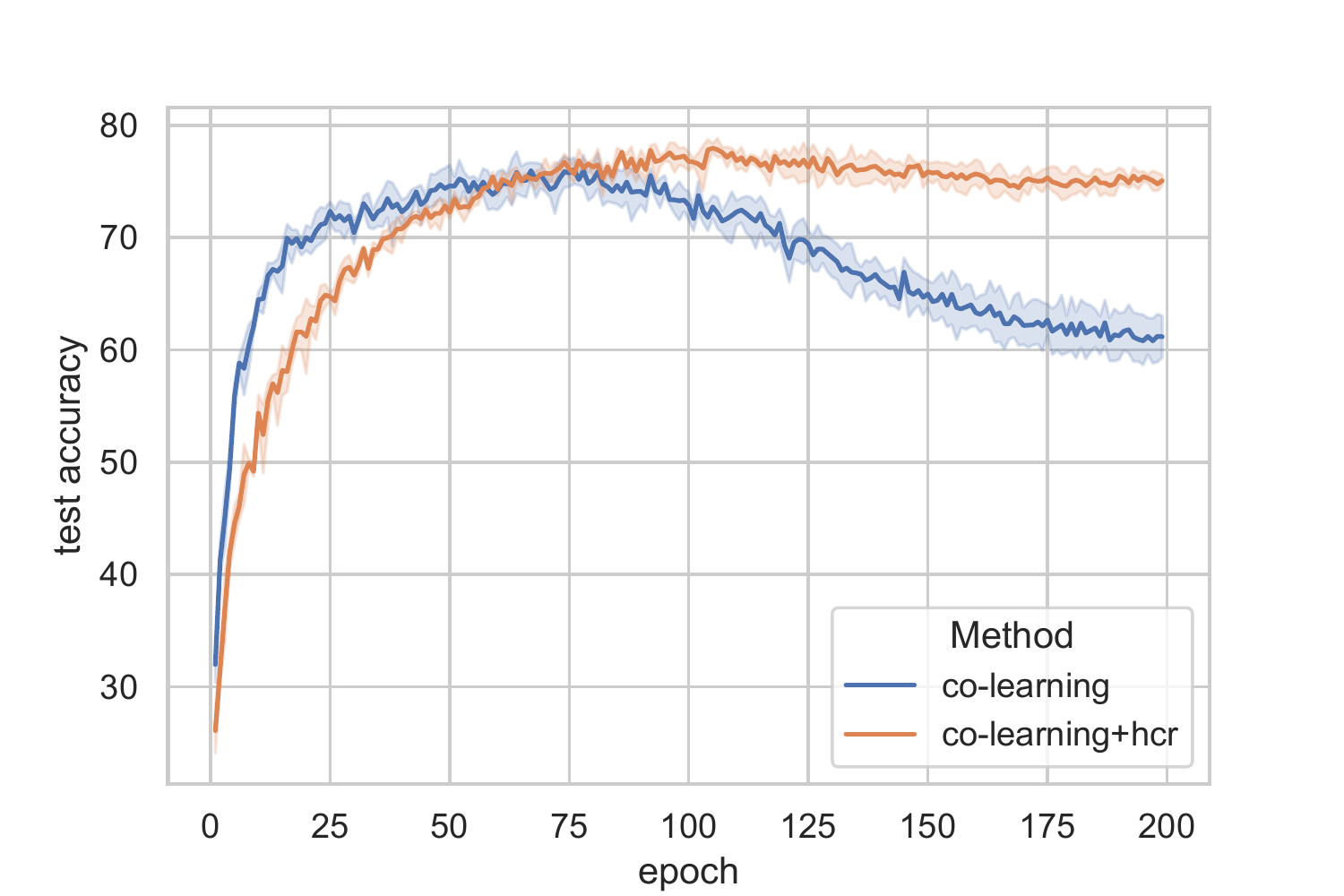}
  \caption{Results on CIFAR-10 with 80\% symmetric noise.}
  \label{fig:cifar10-sym80}
\end{figure}

We also conduct experiments on CIFAR10 with extremely high noise and show the results in Figure \ref{fig:cifar10-sym80}. While Co-learning suffers from overfitting on noisy labels, HCR consistently works well and performs great robustness.

\section{Conclusion}
In this paper, we borrow several theoretical insights from geometry and propose a novel consistency regularization method for semi-supervised and weak-supervised learning, called hyperspherical consistency regularization (HCR), to encourage the pairwise distance distribution of the classifier to be similar to the distribution of the projection head in the latent space. HCR can be conveniently implemented to those jointly learning methods as a plug-in regularization, or be applied to a vanilla supervised learning network with only an additional projection head. Through extensive experiments on semi-supervised learning, fine-grained classification and noisy label learning, HCR shows consistent improvements on these tasks. In general, HCR cast a novel view on leveraging self-supervised learning to assist data-efficient and robust deep learning by introducing hyperspherical consistency.

\paragraph{Acknowledgemnet}

This work is supported in part by the Science and Technology Innovation 2030 - Major Project (No. 2021ZD0150100) and National Natural Science Foundation of China (No. U21A20427).

{\small
\bibliographystyle{ieee_fullname}
\bibliography{ref}
}

\end{document}